\pdfoutput=1

\documentclass[11pt]{article}

\usepackage[]{acl}

\usepackage{times}
\usepackage{CJKutf8}
\usepackage{array}
\usepackage{tcolorbox}
\usepackage{latexsym}
\usepackage{enumitem}
\usepackage{amsmath}
\usepackage{subfigure}
\usepackage{multirow}
\usepackage{booktabs}
\usepackage[T1]{fontenc}

\usepackage[utf8]{inputenc}

\usepackage{microtype}

\usepackage{graphicx}
\newcommand{\modi}{\textcolor{black}}

\definecolor{mygreen}{RGB}{34,139,34}
\newcommand{\good}{\textcolor{mygreen}}
\newcommand{\bad}{\textcolor{red}}
\title{DRT: Deep Reasoning Translation via Long Chain-of-Thought}

\author{Jiaan Wang, \ Fandong Meng\thanks{ \ \ Corresponding author.}, \ Yunlong Liang, \ Jie Zhou \\
Pattern Recognition Center, WeChat AI, Tencent Inc \\ 
\texttt{\{torchwang,fandongmeng,yunlonliang,withtomzhou\}@tencent.com}
}

\begin{document}
\maketitle
\begin{abstract}
Recently, O1-like models have emerged as representative examples, illustrating the effectiveness of long chain-of-thought (CoT) in reasoning tasks such as math and coding tasks.
In this paper, we introduce DRT, an attempt to bring the success of long CoT to neural machine translation (MT).
Specifically, in view of the literature books that might involve similes and metaphors, translating these texts to a target language is very difficult in practice due to cultural differences.
In such cases, literal translation often fails to convey the intended meaning effectively.
Even for professional human translators, considerable thought must be given to preserving semantics throughout the translation process.
To simulate LLMs' long thought ability in MT, we first mine sentences containing similes or metaphors from existing literature books, and then develop a multi-agent framework to translate these sentences via long thought.
In the multi-agent framework, a translator is used to iteratively translate the source sentence under the suggestions provided by an advisor.
To ensure the effectiveness of the long thoughts, an evaluator is also employed to quantify the translation quality in each round.
In this way, we collect tens of thousands of long-thought MT data, which is used to train our DRT.
Using Qwen2.5 and LLama-3.1 as the backbones, DRT models can learn the thought process during machine translation, and outperform vanilla LLMs as well as LLMs which are simply fine-tuning on the paired sentences without long thought, showing its effectiveness.\footnote{The synthesized data and model checkpoints are released at \url{https://github.com/krystalan/DRT}.}

\end{abstract}

\section{Introduction}

Recently, the emergence of the O1-like LLMs shows great performance in reasoning tasks, \emph{e.g.}, math and coding tasks~\cite{openai_o1_2024,qin2024o1,huang2024o1,zhang2024o1,zhao2024marco}.
With the help of long thought, LLMs tend to explore, reflect and self-improve the reasoning processes to achieve more accurate answers.

In this paper, we explore technical routes to bring the success of long thought to MT.
To this end, we introduce DRT, a product of our exploration, and we hope it could facilitate the research community.
There are two key points in achieving this goal:

\romannumeral1) \textbf{A suitable translation scenario to employ long thought in MT}: Not all scenarios require long chain-of-thought (CoT)\footnote{“long CoT” is equal to “long thought”, and we alternatively use these two terms in this paper.} during translation. For example, in simple expressions, literal translation can meet most needs, and translation via long CoT may be unnecessary. Inappropriate scenarios might cause the overthinking issue~\cite{chen2024not}.

\romannumeral2) \textbf{A method to synthesize MT data with long thought}: Long thought SFT (supervised fine-tuning) data plays a vital role in simulating LLMs' long thought ability~\cite{huang2024o1}. Previous work pays much attention to how to synthesize long-thought data in math and coding tasks~\cite{qin2024o1,huang2024o1,zhao2024marco}.

For \romannumeral1), inspired by~\citet{van1981limits}, a possible scenario is translating sentences with similes or metaphors, where literal translation often fails to convey the intended semantics.
Given that, we decide to mine such sentences from literature books. The mining process uses an advanced large language model (LLM) to first judge \texttt{Q1}: \emph{whether each literature sentence has any similes or metaphors}.
If has, the LLM will be asked to literally translate the sentence to a target language, and give a final judgment on \texttt{Q2}: \emph{whether literal translation is effective for native speakers of the target language to comprehend.}
If the answers of \texttt{Q1} and \texttt{Q2} are ``yes'' and ``no'', respectively, the corresponding literature sentences will be reserved, and regarded as ``suitable to translate via long thought''.

For \romannumeral2), after collecting the literal sentences with similes or metaphors, the next question is how to synthesize long thought MT samples.
Previous work typically utilizes Monte Carlo Tree Search (MCTS)~\cite{qin2024o1,zhao2024marco,zhang2024o1} or data distillation~\cite{huang2024o1} (from existing O1-like models) to collect long thought SFT samples.
Nevertheless, MCTS is typically used in math and coding tasks where multiple reasoning behaviors should be considered, and the method emphasizes complex reasoning that might not be efficient for machine translation.
Besides, utilizing existing O1-like models for data distillation might (1) constrain the potential quality of the long-thought data; and (2) have a data gap in MT since current O1-like models are typically optimized toward math and coding tasks.

Therefore, we propose a multi-agent framework to synthesize MT data with long thought.
In detail, there are three agents in the framework, \emph{i.e.}, a translator, an advisor and an evaluator.
The synthesis process is iterative, consisting of the following three steps during each iteration: (1) the translator generates a new translation conditioned on the previous step's translation and the corresponding refinement suggestions from the advisor; (2) the advisor evaluates the current translation and offers detailed feedback; (3) the evaluator assesses the current translation and gives an evaluation score using predefined scoring criteria.
Once the translation score provided by the evaluator reaches a pre-defined threshold or the number of iterations reaches a maximum value, the iteration will stop.
After that, the translation and suggestions in every step could form the long-thought MT samples.
To improve the readability and fluency of the long-thought data, we employ GPT-4o~\cite{hurst2024gpt} to reformulate the long-thought content.

Based on the collected long-thought MT samples, we train our DRT-7B, DRT-8B and DRT-14B using the backbones of Qwen2.5-7B-Instruct, Llama-3.1-8B-Instruct~\cite{dubey2024llama} and Qwen2.5-14B-Instruct~\cite{qwen2.5}, respectively.
Experimental results on literature translation verify their effectiveness.
\modi{In particular, DRT-14B outperforms QwQ-32B-preview and DeepSeek-R1-Distill-Qwen-32B in terms of BLEU, CometKiwi, CometScore and GPT-4 evaluations.
Moreover, human evaluation and case study show the strong translation performance of DRT models.}

Our main contributions are concluded as follows:
\begin{itemize}[leftmargin=*,topsep=0pt]
\setlength{\itemsep}{0pt}
\setlength{\parsep}{0pt}
\setlength{\parskip}{0pt}
\item We propose DRT aiming at building LLMs with long-thought machine translation ability. To achieve this, we mine literature sentences with similes or metaphors, and collect MT samples with long-thought processes.
\item To synthesize the long-thought MT samples, we propose a multi-agent framework that involves a translator, an advisor and an evaluator. These three agents collaborate in an iterative manner to produce long thoughts during MT. Lastly, GPT-4o is used to further improve the quality of the synthesized long-thought MT samples.
\item Experimental results on literature translation verify the effectiveness of our DRT. With the help of long thought, LLMs can learn to think during the machine translation.
\end{itemize}

\section{DRT Data}

We focus on English-to-Chinese translation\footnote{Although we focus on English-to-Chinese translation in this work, the methods we introduced can be trivially applied to other languages or translation directions.}, and we introduce how to collect the long-thought MT samples via three steps in this section:
(1) collecting English sentences that tend to require long thoughts during translation (\S~\ref{subsec:2.1});
(2) synthesizing the long-thought translation process for the collected sentences by a designed multi-agent framework (\S~\ref{subsec:2.2});
(3) improving the readability and fluency of the long-thought content to form the final long-thought MT samples (\S~\ref{subsec:2.3}).
Next, we provide data statistics and data analyses of the collected data to give a deeper understanding (\S~\ref{subsec:2.4}).
Finally, we discuss the data quality (\S~\ref{subsec:2.5}).

\begin{figure*}[t]
\centerline{\includegraphics[width=0.80\textwidth]{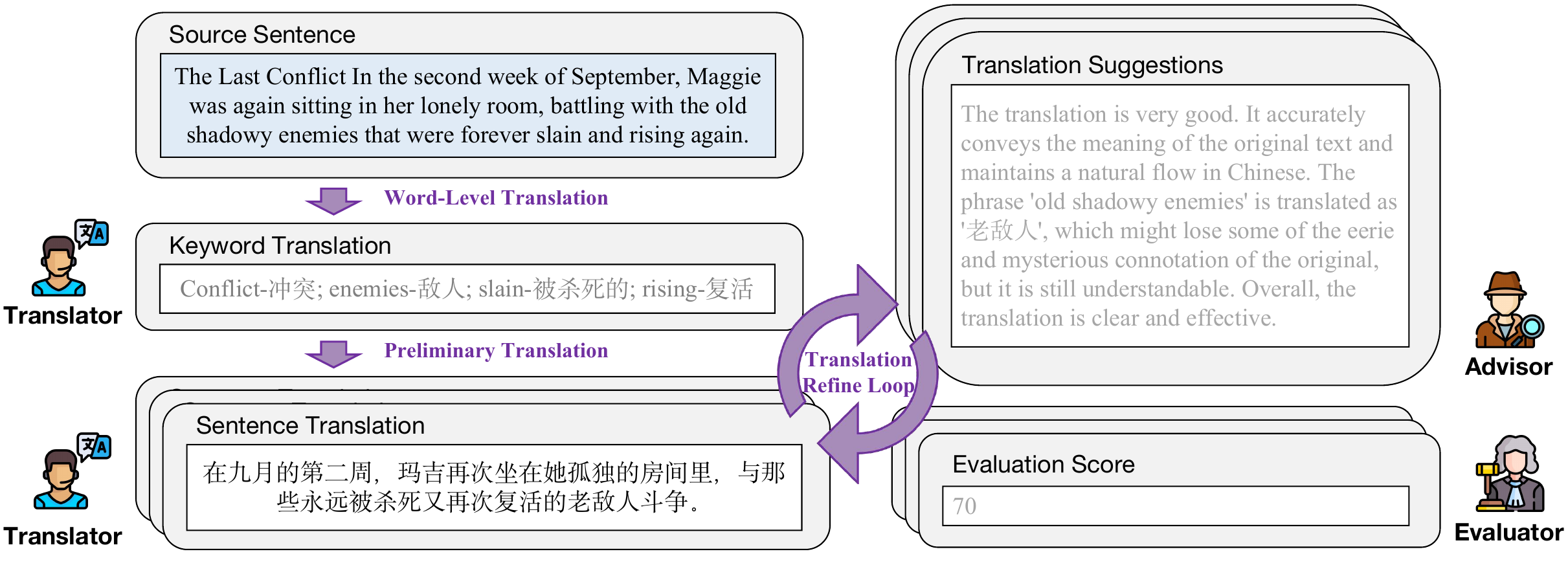}}
\caption{The illustration of the multi-agent framework to synthesize long-thought MT samples. (a) A translator iteratively produces translations under the suggestions provided by an advisor; (b) An advisor reviews the translation results and gives suggestions; (c) An evaluator assesses the translation results and gives an overall score to indicate the translation quality.}
\label{fig:multi_agent_framework}
\end{figure*}

\subsection{Literature Book Mining}
\label{subsec:2.1}

Following~\citet{kryscinski-etal-2022-booksum}, we leverage the literature books from the Project Gutenberg public-domain book repository\footnote{\url{https://www.gutenberg.org/}}, where the books are typically more than fifty years old and their copyrights have expired.
About 400 English books are used to mine sentences with similes or metaphors.

First, we extract all sentences from these books, and filter out too short or too long sentences, \emph{i.e.}, less than 10 words or more than 100 words, resulting in 577.6K literature sentences.
Second, for each sentence, we use Qwen2.5-72B-Instruct~\cite{qwen2.5} to judge whether the sentence involves similes or metaphors, and discard the sentences that do not contain any ones.
Third, for the remaining sentences, we let Qwen2.5-72B-Instruct literally translate them to Chinese, and then judge whether the translation satisfies native Chinese people.
If the answer is negative, the corresponding sentence will be reserved, and regarded as ``suitable to translate via long thought''.
For prompt details, please refer to Appendix~\ref{appendix:prompts_in_literture_book_mining}.
Consequently, we collect 63K (out of 577.6K) literature sentences involving similes or metaphors whose literal translations have flaws, called \emph{pre-collected sentences}.

\subsection{Multi-Agent Framework}
\label{subsec:2.2}

For each pre-collected sentence (denoted as $s$), we design a multi-agent framework to translate it via long thought.
As shown in Figure~\ref{fig:multi_agent_framework}, our multi-agent framework includes three agents: a translator, an advisor, and an evaluator, \modi{each of which use Qwen2.5-72B-Instruct as the backbone.}
The synthetic process is illustrated as follows:

\noindent (1) \emph{Word-level Translation.}
The translator first identifies the keywords that lie in the sentence, and then provides their translations under the consideration of the context.
The keywords are denoted as $\mathcal{W}^{\text{src}} = \{w^{\text{src}}_1, w^{\text{src}}_2, ..., w^{\text{src}}_k\}$, where $w^{\text{src}}_i$ indicates the $i$-th keyword in $s$, and $k$ is the number of keywords.
The translation of keywords is denoted as $\mathcal{W}^{\text{tgt}} = \{w^{\text{tgt}}_1, w^{\text{tgt}}_2, ..., w^{\text{tgt}}_k\}$.
This step enables the model to identify potential challenges in translating the entire sentence by breaking it down into sub-problems (\emph{i.e.}, word-level translation).

\noindent (2) \emph{Preliminary Translation.}
The translator then provides a preliminary sentence translation ($t^0$) conditioned on both the source sentence ($s$) and its keyword bilingual pairs ($\langle \mathcal{W}^{\text{src}},\mathcal{W}^{\text{tgt}} \rangle$).

\noindent (3) \emph{Translation Refine Loop.}
In the refine loop, three agents work together to refine the translation iteratively.
In each iteration step $k$ (start from $k=1$), the advisor first evaluates the translation in the previous step, \emph{i.e.}, $t^{k-1}$, and provides detailed feedback $f^{k-1}$ for polishing it.
Then, the evaluator gives an overall score of $t^{k-1}$ conditioned on both pre-defined scoring criteria and $f^{k-1}$, and the score is denoted as $s^{k-1}$.
In the last of the iteration step, the translator takes its previous translation $t^{k-1}$, the corresponding feedback $f^{k-1}$ and overall score $s^{k-1}$ into account to provide a new translation $t^k$.
The translation refine loop will stop when the overall score reaches a pre-defined threshold or the number of iteration steps meets the maximum.
For prompt details of the translator, advisor and evaluator, please refer to Appendix~\ref{appendix:prompts_in_multi_agent_framework}.

\begin{figure*}[t]
\centerline{\includegraphics[width=0.80\textwidth]{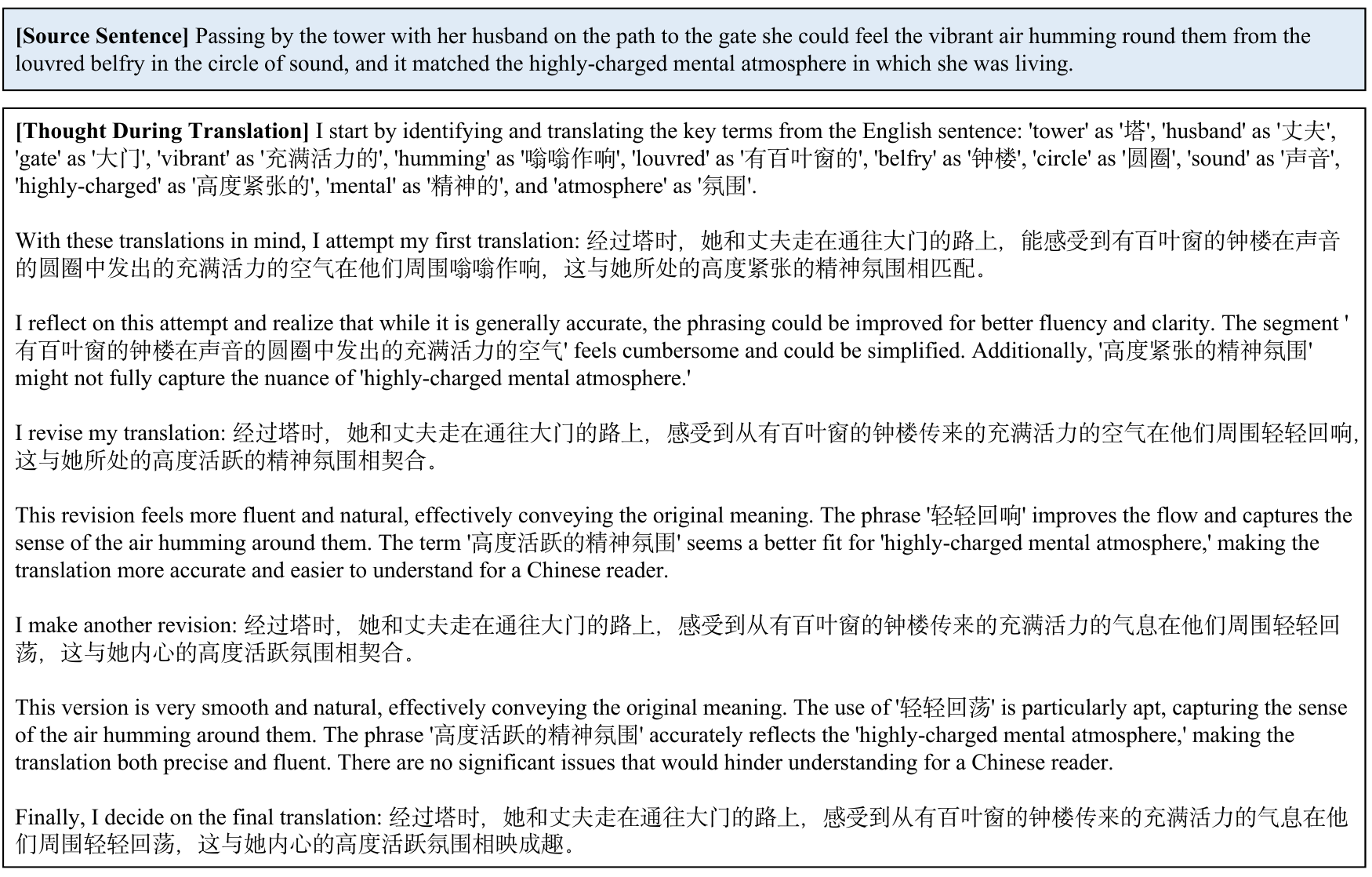}}
\caption{An example of long thought synthesized by the designed multi-agent framework and GPT-4o reformulation.}
\label{fig:data_case}
\end{figure*}

\subsection{Long Thought Reformulation}
\label{subsec:2.3}

After the multi-agent collaboration, we obtain a long thought process:
\begin{equation}
\begin{split}
\mathcal{P}(s): s \Rightarrow \langle \mathcal{W}^{\text{src}},\mathcal{W}^{\text{tgt}} \rangle \Rightarrow \langle t^0, f^0, s^0 \rangle \\
\Rightarrow \langle t^1, f^1, s^1 \rangle \Rightarrow ... \Rightarrow \langle t^m, f^m, s^m \rangle
\end{split}
\end{equation}
where $\mathcal{P}(s)$ denotes the multi-agent thought process for $s$, and $m$ is the number of iteration steps.
To emphasize the valid thought process, translations without score change will be removed.
That is, if $s^i$ is equal to $s^{i-1}$ ($i=1,2,...,m$), we will discard $\langle t^i, f^i, s^i \rangle$ in $\mathcal{P}(s)$, resulting in:
\begin{equation}
\begin{split}
\mathcal{P}^{'}(s): s \Rightarrow \langle \mathcal{W}^{\text{src}},\mathcal{W}^{\text{tgt}} \rangle \Rightarrow \langle t^0, f^0, s^0 \rangle \\
\Rightarrow \langle t^{r_1}, f^{r_1}, s^{r_1} \rangle \Rightarrow ... \Rightarrow \langle t^{r_n}, f^{r_n}, s^{r_n} \rangle
\end{split}
\end{equation}
where $1 \leq r_1 < r_2 < ... < r_n \leq m$, and $n$ is the number of remaining steps.
If $n < 3$, we will discard the whole sample, \emph{i.e.}, $\mathcal{P}(s)$.

For the remaining samples, we follow \citet{qin2024o1}, and leverage GPT-4o~\cite{hurst2024gpt} to modify and polish $\mathcal{P}^{'}(s)$ into a self-reflection description (the used prompt is provided in Appendix~\ref{appendix:prompts_in_thought_reformulation}).
Finally, we obtain 22,264 MT samples with long thought.
Figure~\ref{fig:data_case} gives an example sample to illustrate the synthetic results.

\begin{table}[t]
\centering
\resizebox{0.45\textwidth}{!}
{
\begin{tabular}{lcccc}
\toprule[1pt]
          & \# Sample & Query & Thought & Output \\ \midrule[1pt]
o1-journey & 327       & 41.53 & 486.05  & 3.41   \\
Marco-O1 CoT data     & 10,000       & 52.73 & 673.98  & 52.73  \\ \midrule
DRT data $_\text{(training)}$     & 19,264     & 37.25 & 527.64  & 44.67  \\
DRT data $_\text{(validation)}$       & 1,000      & 37.43 & 531.36  & 44.98  \\
DRT data $_\text{(testing)}$       & 2,000      & 37.19 & 525.44  & 44.70  \\ \bottomrule[1pt]
\end{tabular}
}
\caption{The number of samples and average token-level length of query, thought and output. ``Query'' and ``Output'' in DRT data mean the source sentences and the translated outputs, respectively.} 
\label{table:data_statistics}
\end{table}

It is also worth noting that during the GPT-4o reformulation, we specify the translation with the highest score $s^{r_j}$ as the final translation. Thus, the final translation is not necessarily the last one during refinement, \emph{i.e.}, $t^{r_n}$.

\begin{figure}[t]
\centerline{\includegraphics[width=0.40\textwidth]{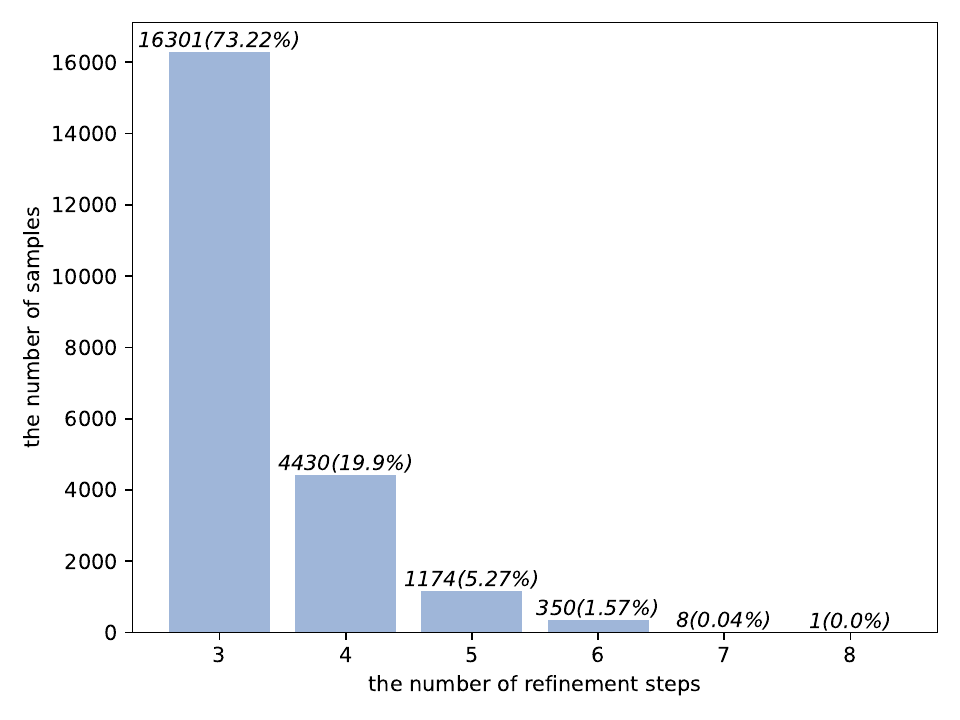}}
\caption{The distribution of the number of refinement steps in DRT data.}
\label{fig:refine_num}
\end{figure}

\subsection{Data Statistics and Data Analyses}
\label{subsec:2.4}

\begin{figure*}[t]
\centering
\subfigure[]{
  \includegraphics[width=0.233\linewidth]{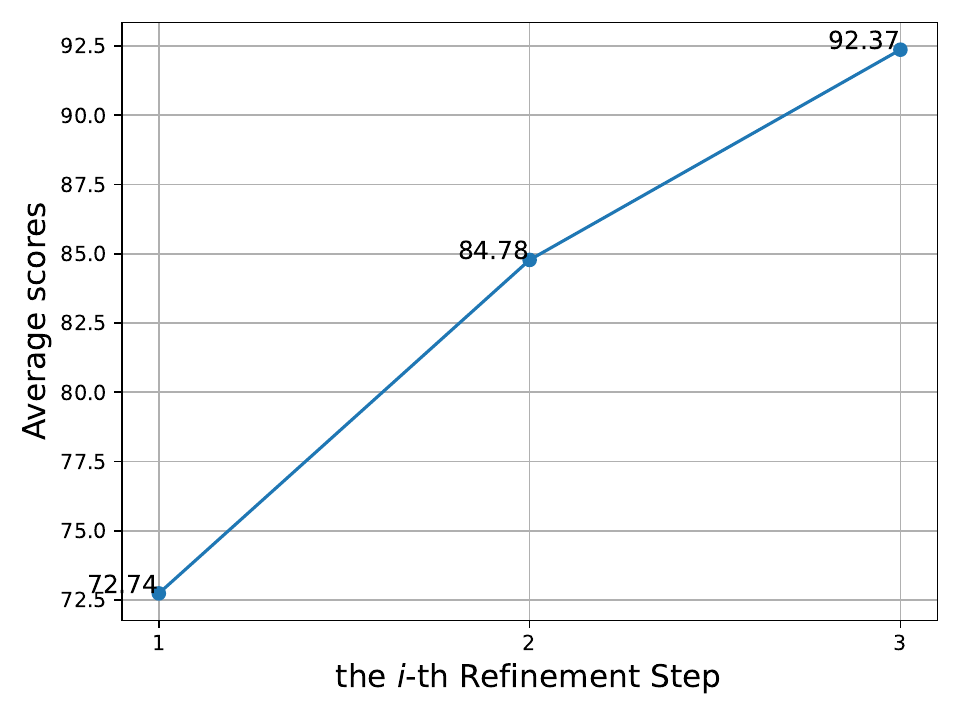}
}
\subfigure[]{
  \includegraphics[width=0.233\linewidth]{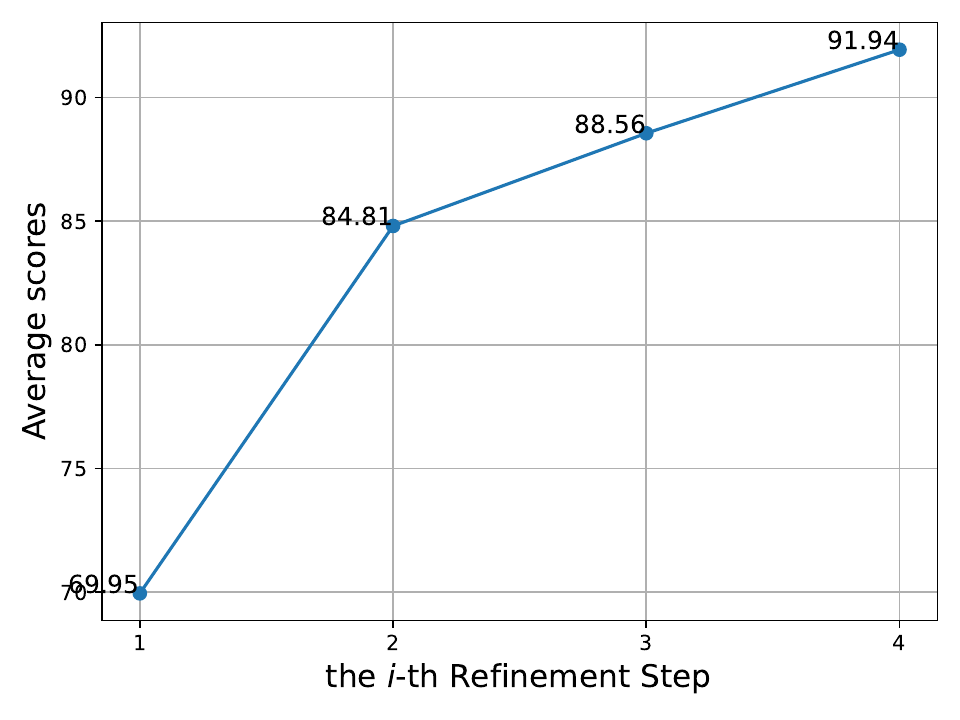}
}
\subfigure[]{
  \includegraphics[width=0.233\linewidth]{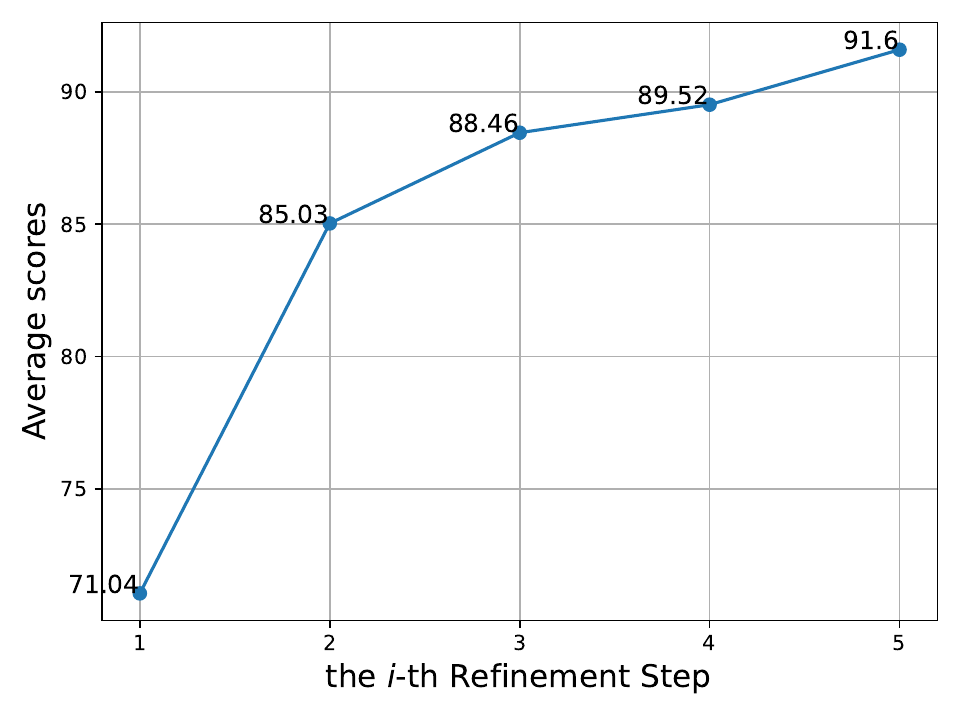}
}
\subfigure[]{
  \includegraphics[width=0.233\linewidth]{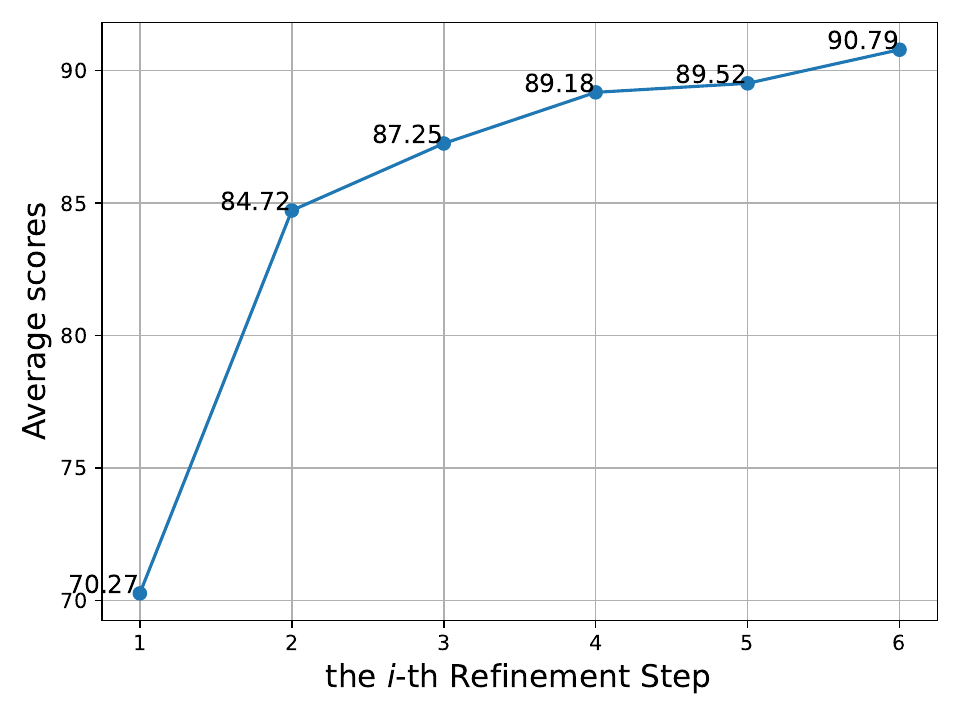}
}
\caption{\modi{Trends in average scores (provided by the evaluator agent) over the refinement steps. The trends for samples with three, four, five, and six refinement steps are illustrated in (a), (b), (c), and (d), respectively.}}
\label{fig:score_change}
\end{figure*}

We split the collected 22,264 samples into training, validation and testing sets with 19,264, 1,000 and 2,000 samples, respectively.
Table~\ref{table:data_statistics} shows the data statistics of DRT data and previous O1-like data.
For Marco-O1 CoT data~\cite{zhao2024marco}, since it is not fully released, we use its demo data to calculate the data statistics.\footnote{\url{https://github.com/AIDC-AI/Marco-o1}}
As we can see, the average number of tokens in our synthesized thought reaches 500+ tokens, showing the long thought process in our data.

\noindent
\emph{Refine Loop Analyses.}
Figure~\ref{fig:refine_num} shows the number of refinement steps in the DRT data, which ranges from 3 to 8 steps.
We can find that most samples (73.22\%) involve 3 refinement steps, while only one sample involves 8 steps.
Furthermore, to provide a deeper understanding of the refinement process, we calculate the average edit distance before and after each refinement step.
Specifically, the first three refinement steps cause 21.44, 13.16 and 10.90 character-level edit distance.
This observation is consistent with intuition. As the refinement progresses, the magnitude of the modification gradually decreases.
\modi{To further understand the improvement brought by the translation refine loop, we calculate the average overall scores (provided by the evaluator agent) along with each refinement step.
As shown in Figure~\ref{fig:score_change}, as the number of refinement steps increases, the average score generally increases, demonstrating that the refine loop could iteratively increase the quality of translations.}

\vspace{0.2ex}

\begin{table}[t]
\centering
\resizebox{0.42\textwidth}{!}
{
\begin{tabular}{lccc}
\toprule[1pt]
\multicolumn{1}{c}{Metric}                & ACC. (\%)  \\ \midrule[1pt]
CometKiwi       & 56.0 \\
Evaluator Agent (Qwen2.5-72B-Instruct) & 92.5  \\
Evaluator Agent (GPT-4o)          & 93.5  \\ \bottomrule[1pt]
\end{tabular}
}
\caption{\modi{Accuracy of automatic metrics for translation quality estimation (ACC.: accuracy).}}
\label{table:correlation_of_evaluator}
\end{table}

\subsection{Quality Analyses}
\label{subsec:2.5}

\noindent \modi{\emph{The Effectiveness of the Evaluator Agent.}}
\modi{Previous work has shown that the state-of-the-art LLMs can be used as evaluators for various text generation tasks~\cite{kocmi2023large,wang-etal-2023-chatgpt,li-etal-2024-leveraging-large}.
To figure out the effectiveness of our evaluator agent, we randomly select 200 source sentences from DRT data, and for each of them, we further select its two translations as well as scores (provided by the evaluator agent) during refinement.
We next employ human annotators to compare the two translations of each source sentence, and judge which translation is better, or two translations are similar in quality (annotation details can be found in Appendix~\ref{appendix:human_annotation}).
After obtaining the quality labels, we calculate the accuracy of the evaluator agent according to its evaluation score.
For comparison, we also calculate the accuracy of CometKiwi~\cite{rei-etal-2022-cometkiwi} and GPT-4o evaluator agent.
As shown in Table~\ref{table:correlation_of_evaluator}, our evaluator agent achieves a high accuracy (92.5\%), demonstrating its effectiveness in evaluating literature translation quality.
Besides, the widely-used CometKiwi metric only achieves 56.0\% accuracy. Thought CometKiwi is powerful in the general domain (\emph{e.g.}, news)~\cite{kocmi2023large}, its effectiveness in the literature domain is limited and unreliable, which is also pointed out by \citet{karpinska-iyyer-2023-large}.
Furthermore, the GPT-4o evaluator agent slightly outperforms the origin evaluator agent (with Qwen2.5-72B-Instruct backbone).
Considering the tradeoff between cost and effectiveness, we finally decide to use Qwen2.5-72B-Instruct as our evaluator agent.
}

\vspace{0.2ex}
\noindent \modi{\emph{Translation Quality.}}
\modi{Based on the effectiveness of the evaluator agent and the observation that evaluation scores of final translations typically reach 90.0 (c.f., Figure~\ref{fig:score_change}), we can ensure a high level of translation quality in the constructed data. According to the pre-defined scoring criteria of the evaluator agent (c.f., Appendix~\ref{appendix:prompts_in_multi_agent_framework}), a score of 90.0 indicates excellent translations.}

\begin{table*}[t]
\centering
\resizebox{0.75\textwidth}{!}
{
\begin{tabular}{lcccccc}
\toprule[1pt]
\multicolumn{1}{c}{\multirow{2}{*}{Model}} & \multicolumn{3}{c}{reference-free} & \multicolumn{3}{c}{reference-based}                        \\
\cmidrule(r){2-4} \cmidrule(r){5-7} \multicolumn{1}{c}{}                       & GEA &  GRF               & CometKiwi             & GRB   & \multicolumn{1}{c}{BLEU} & \multicolumn{1}{c}{CometScore} \\ \midrule[1pt]
\multicolumn{7}{c}{\emph{Vanilla LLMs}} \\ \midrule
Llama-3.1-8B-Instruct & 59.58  & 79.25  & 70.14  &  73.30 & 18.55  & 74.58 \\
Qwen2.5-7B-Instruct   & 66.21 & 81.53  & 70.36  & 77.92  & 27.02     & 76.78      \\
Qwen2.5-14B-Instruct & 70.86  & 84.74 & 72.01    & 80.85  & 30.23    & 78.84      \\
Marco-o1-7B  & 64.24   & 82.41 & 71.62  &  77.50 & 29.48     & 77.41      \\ 
QwQ-32B-preview & \underline{75.50}  & \underline{86.31} & 71.48  &  \underline{83.08} & 27.46     & 78.68      \\
DeepSeek-R1-Distill-Llama-8B      &  56.89     &    76.31     &    67.13     &    69.49     &    15.83     &    71.82 \\
DeepSeek-R1-Distill-Qwen-7B      &     43.66     &    65.16     &    63.49     &    58.13     &    10.99     &    69.21 \\
DeepSeek-R1-Distill-Qwen-14B      &     70.64     &    83.92     &    71.01     &    80.29     &    25.55     &    77.66 \\
DeepSeek-R1-Distill-Qwen-32B      &     71.88     &    84.78     &    71.93     &    81.59     &    29.36     &    78.93 \\ \midrule[1pt]
\multicolumn{7}{c}{\emph{SFT LLMs (w/o CoT)}} \\ \midrule
Llama-3.1-8B-SFT & 69.33  & 84.10 & 70.25  &  80.18  & 30.03   & 78.26 \\ 
Qwen2.5-7B-SFT  & 72.29  & 85.06 &  71.03 & 81.72  & 35.44  &  80.10  \\
Qwen2.5-14B-SFT  & 74.53   & 85.66 &  \underline{72.08}  &  \underline{83.08} & \textbf{37.63}  &  \textbf{80.82}    \\ \midrule[1pt]
\multicolumn{7}{c}{\emph{DRT}} \\ \midrule
DRT-8B $_\text{(Backbone: Llama-3.1-8B-Instruct)}$  & 69.65$^{\dagger}$  & 84.49$^{\ddagger}$ & 70.85$^{\dagger}$ & 80.80$^{\dagger}$  & 32.67$^{\dagger}$  & 78.81$^{\dagger}$ \\ 
DRT-7B $_\text{(Backbone: Qwen2.5-7B-Instruct)}$  & 75.05$^{\dagger}$  & 85.57$^{\ddagger}$ &  71.78$^{\dagger}$    & 82.38$^{\dagger}$  & 35.54   & 80.19$^{\ddagger}$    \\
DRT-14B $_\text{(Backbone: Qwen2.5-14B-Instruct)}$  & \textbf{77.41}$^{\dagger}$  & \textbf{87.19}$^{\dagger}$  & \textbf{72.11}   &  \textbf{83.20}$^{\ddagger}$ & \underline{36.46}     & \underline{80.64}     \\ \bottomrule[1pt]
\end{tabular}
}
\caption{Experimental results on literature translation. The \textbf{bold} and the \underline{underline} denote the best and second-best performances, respectively. \modi{``$\dagger$'' and``$\ddagger$'' denote statistically significant better than the corresponding SFT LLMs (w/o CoT) with t-test p < 0.01 and 0.05, respectively.}} 
\label{table:main_res}
\end{table*}

\section{Experiments}

\subsection{Experimental Setups}

\noindent \textbf{Metrics.}
Following previous work, we adopt \emph{``BLEU''}~\cite{papineni-etal-2002-bleu}, \emph{``CometKiwi''} and \emph{``CometScore''}~\cite{rei-etal-2022-cometkiwi} to evaluate the model translations.
Among them, BLEU evaluates n-grams overlap between model translations and references, while CometScore evaluates the semantic similarity of model translations against references.
CometKiwi uses a language model to judge whether a model translation conveys the semantics of the source sentence.

\modi{As pointed out by \citet{karpinska-iyyer-2023-large}, BLEU and COMET may be ineffective for evaluating literature translation.}
Meanwhile, recent studies also show the strong ability of LLMs in NLP evaluation~\cite{li-etal-2024-leveraging-large}.
Therefore, we use evaluators implemented using GPT-4o in reference-based and reference-free styles, which we refer to as \emph{``GRB''} and \emph{``GRF''}, respectively.
The evaluation prompts borrow from \citet{kocmi2023large}, and are illustrated in Appendix~\ref{appendix:gpt_4o_evaluator}.
\modi{Furthermore, as demonstrated in \S~\ref{subsec:2.4}, the GPT-4o evaluator agent achieves great accuracy in literature translation.
We also leverage it as the evaluation metric in experiments, which is referred to as \emph{``GEA''}.
Since GRB, GRF and GEA need the API costs, we randomly select 400 samples to conduct evaluation.}

\vspace{0.2ex}
\noindent \textbf{Backbones.}
We adopt the following three LLMs as the backbones of our DRT: Llama-3.1-8B-Instruct~\cite{dubey2024llama}, Qwen2.5-7B-Instruct and Qwen2.5-14B-Instruct~\cite{yang2024qwen2}.
All model checkpoints are publicly available.

For evaluation toolkits and the implementation details of all models, please refer to Appendix~\ref{appendix:implementation_details}.

\subsection{Comparison Models}

\noindent \emph{Vanilla LLMs.}
We leverage vanilla Llama-3.1-8B-Instruct, Qwen2.5-7B-Instruct and Qwen2.5-14B-Instruct~\cite{qwen2.5} as the comparison models.
Besides, six O1-like LLMs are also conducted as baselines: Marco-o1-7B~\cite{zhao2024marco}, QwQ-32B-preview~\cite{team2024qwq}, \modi{DeepSeek-R1-Distill-Qwen-7B, DeepSeek-R1-Distill-Llama-8B, DeepSeek-R1-Distill-Qwen-14B and DeepSeek-R1-Distill-Qwen-32B~\cite{guo2025deepseek}.}

\vspace{0.5ex}
\noindent \emph{SFT LLMs (w/o CoT).}
We also fine-tune LLMs with only paired sentences of DRT training data (without thought).
This setting allows LLMs to learn the mapping from source literature sentences to the corresponding Chinese translations directly.
We denote the fine-tuned LLMs as Llama-3.1-8B-SFT, Qwen2.5-7B-SFT and Qwen2.5-14B-SFT, serving as strong baselines in the experiments.

\subsection{Main Results}

Table~\ref{table:main_res} shows the experimental results, we analyze the performance from the following aspects:

\vspace{0.5ex}
\noindent \textbf{SFT LLMs (w/o CoT) vs. Vanilla LLMs.}
After instruction tuning on the paired sentences of our training data, SFT LLMs (w/o CoT) significantly outperform the corresponding vanilla LLMs.
For example, Llama-3.1-8B-SFT outperforms Llama-3.1-8B-Instruct by 9.75 GEA, 4.85 GRF and 6.88 GRB. Qwen2.5-7B-SFT outperforms Qwen2.5-7B-Instruct by 6.08 GEA, 3.53 GRF and 3.80 GRB.
This finding demonstrates the effectiveness of our multi-agent framework and the quality of the synthesized translation.
Please also note that the final translations are synthesized by Qwen2.5-72B-Instruct, indicating that we can leverage off-the-shelf \emph{open-source} LLMs to collect high-quality literation translation data. And the data could help smaller LLMs (such as 7B and 14B ones) to boost their literature translation skills.

\vspace{0.5ex}
\noindent \textbf{DRT vs. Vanilla LLMs.}
After fine-tuning on the long-thought MT training data, our DRT-series LLMs also significantly outperform the corresponding vanilla backbones.
Particularly, DRT-14B outperforms QwQ-32B-preview and DeepSeek-R1-Distill-Qwen-32B in terms of all metrics, showing its effectiveness in literature MT.

\vspace{0.5ex}
\noindent \textbf{DRT vs. SFT LLMs (w/o CoT).}
Using Llama-3.1-8B-Instruct and Qwen2.5-7B-Instruct as backbones, LLMs tuned with long thought achieve better performance than those tuned without long thought in terms of all metrics.
For example, DRT-7B outperforms Qwen2.5-7B-SFT by 2.76 GEA, 0.51 GRF, 0.75 CometKiwi, 0.66 GRB, 0.10 BLEU and 0.09 CometScore.
When using Qwen2.5-14B-Instruct as the backbone, we find that DRT-14B outperforms Qwen2.5-14B-SFT in terms of GEA, GRF, CometKiwi and GRB, but underperforms in terms of BLEU and CometScore.
In detail, BLEU and CometScore evaluate the translations from the perspective of similarity between model translations and golden references.
We conjecture that the higher BLEU and CometScore performance of Qwen2.5-14B-SFT is due to the model's ability to quickly learn domain-specific translations through tuning without long thoughts, allowing it to adapt to the literature translation more straightforwardly.
However, training without long thoughts might lead the model to a sub-optimal solution, like learning shortcuts.
When adopting evaluation metrics that are not significantly dependent on the golden references (\emph{i.e.}, GEA, GRF, CometKiwi and GRB), DRT-14B shows its superior performance.
Note that although GRB is a reference-based metric, it does not assess the model translations simply based on how similar they are to the golden references.

\vspace{0.5ex}
\noindent \textbf{DRT vs. Commercial LLMs.} \modi{To give a deeper understanding of our DRT models' performance, we further compare DRT models with GPT-4o~\cite{hurst2024gpt} and o1-preview~\cite{openai_o1_2024}. The experimental results and corresponding analyses are provided in Appendix~\ref{appendix:commercial_llms}.}

\begin{table}[t]
\centering
\resizebox{0.38\textwidth}{!}
{
\begin{tabular}{lccc}
\toprule[1pt]
\multicolumn{1}{c}{Model} & Flu.   & Sem.   & Lit.   \\ \midrule[1pt]
Qwen2.5-14B-Instruct      & -0.353 & -0.363 & -0.442 \\
QwQ-32B-Preview           & -0.063 & 0.022  & -0.007 \\
Qwen2.5-14B-SFT           & 0.103  & 0.108  & 0.087  \\
DRT-14B                   & \textbf{0.313}  & \textbf{0.233}  & \textbf{0.362}  \\ \bottomrule[1pt]
\end{tabular}
}
\caption{\modi{Human evaluation results in terms of fluency, semantic accuracy and literary quality.}}
\label{table:human_eval}
\end{table}

\subsection{Human Evaluation}

\modi{We conduct human evaluation to further evaluate the performance of DRT-14B and strong baselines (Qwen2.5-14B-Instruct, QwQ-32B-Preview and Qwen2.5-14B-SFT).
We randomly select 200 samples from our test set, and employ three human evaluators with high levels of fluency in English and Chinese to assess the generated translations from three aspects: fluency (Flu.), semantic accuracy (Sem.) and literary quality (Lit.).
Following the Best-Worst Scaling method~\cite{kiritchenko-mohammad-2017-best}, evaluators are asked to select the best and the worst generated translation on each aspect.
The result scores are calculated based on the percentage of times each model is selected as best minus the times it is selected as worst. Thus, the final scores should range from -1 (worst) to 1 (best).
As shown in Table~\ref{table:human_eval}, DRT-14B significantly outperforms these strong baselines, especially in the aspect of literary quality. These results demonstrate the superiority of our DRT models.
The Fleiss’ Kappa scores~\cite{fleiss1971measuring} of Flu., Sem. and
Lit. are 0.75, 0.69 and 0.85, respectively, indicating a good inter-agreement among evaluators.
}

\subsection{Inference Time Analysis}

\begin{figure}[t]
\centerline{\includegraphics[width=0.35\textwidth]{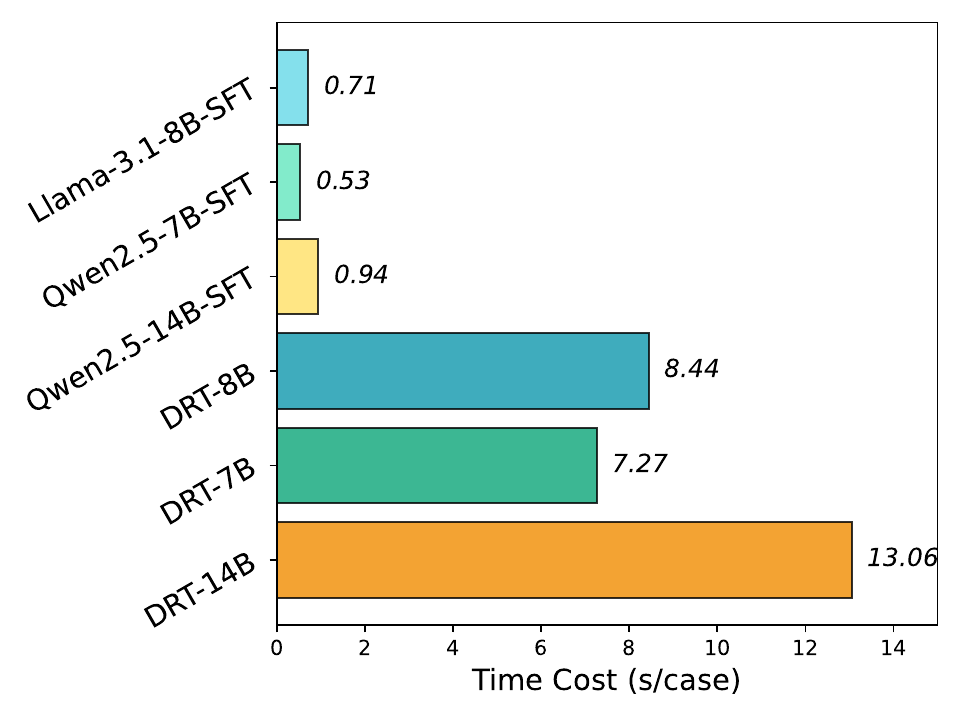}}
\caption{Time cost during inference on the testing set.}
\label{fig:time_cost}
\end{figure}

During evaluating LLMs' literature translation performance on our test set, we leverage vLLM to accelerate the model generation.
A single NVIDIA A100 GPU (40G) is used to deploy each LLM.
As shown in Figure~\ref{fig:time_cost}, the average time costs of DRT models are significantly higher than LLMs (w/o CoT).
This is because DRT models should first generate the long thought and then provide the final translation, thus needing more inference time ($\times$11.9\textasciitilde13.9).
This also indicates that the O1-like LLMs may not be applicable to some scenarios with high real-time requirements.

\begin{table*}[t]
\centering
\resizebox{0.98\textwidth}{!}
{
\begin{tabular}{p{5cm}p{5cm}p{5cm}p{5cm}p{5cm}}
\toprule[1pt]
\multicolumn{1}{c}{Source Sentence}   & \multicolumn{1}{c}{Qwen2.5-14B-Instruct}    & \multicolumn{1}{c}{QwQ-32B-Preview}   & \multicolumn{1}{c}{Qwen2.5-14B-SFT}     & \multicolumn{1}{c}{DRT-14B}     \\ \midrule[1pt]
And if their words had taken another turn...if he himself had been less fastidious about intruding on another man's secrets...it was cruel to think how thin a film had shut out rescue from all this guilt and misery. & \begin{CJK}{UTF8}{gbsn} 而如果他们的言辞\bad{转向了另一个方向}……如果他自己对\bad{侵扰他人秘密}不那么介意……想到正是那层\good{薄薄的隔阂}将救援拒之门外，使得这一切的罪恶和痛苦无法避免，真是残酷。\end{CJK} & \begin{CJK}{UTF8}{gbsn} 而且，\bad{假如他们的谈话走向不同}……假如他自己不对干涉他人私事如此谨慎……那时他会意识到，只要一层薄薄的屏障就阻隔了救援，使得所有这些罪恶和痛苦无法解脱，这真是残忍。\end{CJK} & \begin{CJK}{UTF8}{gbsn} \bad{如果}他们的言辞稍有不同，\bad{若}他不那么介意侵扰别人的隐私，想到那层薄纱如何隔绝了救援，真是残忍。\end{CJK}  & \begin{CJK}{UTF8}{gbsn} \good{若他们的言辞稍作转变}……若他本人不那么介意\good{侵入他人的心扉}……想到这层\good{薄薄的隔阂}竟将所有内疚与痛苦隔离于救援之外，实为残忍。\end{CJK} \\ \midrule
This cold officer upon a monument, who dropped epithets unconcernedly down, would be finer as a dead man, he thought.                                                                                                     & \begin{CJK}{UTF8}{gbsn} 他认为，这个站在纪念碑上的冷漠官员，\bad{若死了会更好}，\bad{他不带任何感情地抛下了一些称呼}。   \end{CJK}                                  & \begin{CJK}{UTF8}{gbsn} 这个冷冰冰的官员站在纪念碑上，\good{毫不在意}地抛下一些称号，他想，\bad{如果作为一个死人会更出色}。   \end{CJK}                                    & \begin{CJK}{UTF8}{gbsn} 这位冷酷的官员站在纪念碑上，\bad{随意地抛出贬义词}，他觉得，\bad{作为死人会更显尊贵}。    \end{CJK}       & \begin{CJK}{UTF8}{gbsn} 纪念碑上的冷淡官员，\good{漫不经心地吟咏那些修饰语}，他心想，若\good{化为亡者}，或许更显尊贵。   \end{CJK}     \\ \midrule
I see on your cheek two tears which I know are hot as two sparks, and salt as two crystals of the sea. & \begin{CJK}{UTF8}{gbsn} 我看见你脸颊上有两滴泪水，我知道它们\good{像两簇火花一样滚烫}，像两颗海水结晶一样咸。 \end{CJK} & \begin{CJK}{UTF8}{gbsn} 我见你面颊两颗泪，\good{温热如火花}，\good{咸涩如海晶}。 \end{CJK} & \begin{CJK}{UTF8}{gbsn} 我见你脸颊上两颗泪珠，热如火花，咸似海盐。 \end{CJK} & \begin{CJK}{UTF8}{gbsn} 我看见你脸颊上的两滴眼泪，\good{它们既如两簇火花般炽热}，\good{又似两粒海盐般咸涩}。 \end{CJK} \\ \midrule
The night wore out, and, as he stood upon the bridge listening to the water as it splashed the river-walls of the Island of Paris, where the picturesque confusion of houses and cathedral shone bright in the light of the moon, the day came coldly, looking like a dead face out of the sky. & \begin{CJK}{UTF8}{gbsn} 夜深了，当他站在桥上，听着水花拍打巴黎岛河岸的声音，月光下房屋与教堂\good{错落有致}的景象闪耀着光芒时，冷冰冰的白天来临了，\bad{仿佛是从天空中露出的一张死人脸}。 \end{CJK} & \begin{CJK}{UTF8}{gbsn} 夜晚消逝了，当他站在桥上，\good{聆听}水流拍打着巴黎岛的河岸，那里的房屋与大教堂构成了一幅\bad{混乱而又美丽}的图景，在月光下闪耀着明亮的光芒，而白天则冷酷地到来，仿佛是\bad{天空中一张死人脸}。 \end{CJK} & \begin{CJK}{UTF8}{gbsn} \good{夜幕渐渐消散}，他站在桥上，\good{聆听}水花轻拍巴黎岛的河岸。月光下，岛上房屋与大教堂\good{交织成一幅迷人的画卷}，显得格外明亮。\good{清晨悄然而至}，\good{宛如从天际浮现的一张冷峻的脸庞。} \end{CJK} & \begin{CJK}{UTF8}{gbsn} 夜色渐渐消逝，他站在桥上，\good{倾听}河水轻拍巴黎城岛的河岸。月光下，房屋与大教堂\good{错落交织}，\good{闪耀着迷人的光辉}。白日冷冰冰地降临，\good{宛如从天际浮现的一张死寂的}\bad{脸}。 \end{CJK} \\  \bottomrule[1pt]       
\end{tabular}
}
\caption{Case Studies of literature translation. \good{Green} indicates good translations, while \bad{red} indicates bad ones.} 
\label{table:case_study}
\end{table*}

\subsection{Case Study}

Table~\ref{table:case_study} provides some literature translation cases of Qwen2.5-14B-Instruct, QwQ-32B-Preview, Qwen2.5-14B-SFT and DRT-14B.
With the help of long thought, the translations of DRT-14B align more closely with the conventions of the Chinese language and exhibit a greater literary quality.
In addition to DRT-14B, some translation snippets of other LLMs can also show a great performance (marked in green).
This indicates that vanilla LLMs might have the capability to translate literature, and long thought could further activate this capability.

\section{Related Work}

\noindent \textbf{O1-like LLMs.}
Recently, O1-like LLMs have shown great performance in reasoning tasks, especially math and coding tasks.
After the emergency of OpenAI O1 model~\cite{openai_o1_2024}, many efforts are given in reproducing OpenAI O1.
For example, \citet{qin2024o1} propose journey learning, a training paradigm, to encourage LLMs to learn not just shortcuts, but the complete exploration process.
%
\citet{huang2024o1} explore the data distillation from existing O1-like models, and show the effectiveness of data distillation.
\citet{zhang2024o1} leverage Monte Carlo Tree Search (MCTS) to synthesize reasoning-enhanced code data, and train O1-Coder.
Marco-o1~\cite{zhao2024marco} is proposed to deal with open-ended text generation.
\modi{More recently, DeepSeek-R1~\cite{guo2025deepseek} and Kimi K1.5~\cite{team2025kimi} are proposed, and show their promising reasoning ability.}

\vspace{0.2ex}
\noindent \textbf{Literature Translation.}
Different from translating standard MT corpora (\emph{e.g.}, news articles), translating literature books is more difficult since it often requires equivalence beyond the word level~\cite{thai-etal-2022-exploring}.
Besides, it is also difficult to evaluate literature translation using automatic metrics, and previous literature translation work typically relies on human evaluation~\cite{fonteyne-etal-2020-literary,karpinska-iyyer-2023-large}.
Due to its difficulty, early work is limited to small-scale attempts~\cite{genzel2010poetic,jones2013faithful,besacier2015automated,toral2018post}.
Recently, \citet{karpinska-iyyer-2023-large} utilize LLMs to perform literature translation, and show that discourse-level LLM translators achieve better performances compared with sentence-level approaches.
\citet{thai-etal-2022-exploring} introduce Par3 to benchmark LLMs' literature translation capability from non-English languages to English.

\section{Conclusion}

In this paper, we introduce DRT, an attempt to bring the success of long-thought reasoning to neural machine translation (MT).
Specifically, we synthesize the machine translation long-thought samples by a designed multi-agent framework and GPT-4o reformulation.
To collect the source sentences that are suitable for translation via long thought, we mine sentences with similes or metaphors from existing literature books.
To synthesize the long thought machine translation process for these sentences, a translator, an advisor and an evaluator collaborate to translate the source sentence iteratively.
Based on the synthesized data, we train DRT models. Extensive experiments on literature translation demonstrate the effectiveness of DRT models in terms of automatic evaluation. Case study and human evaluation further verify their superiority.

\section*{Limitations}

While we show the effectiveness of long thought in MT, there are some limitations worth noting:
(1) We focus on English-to-Chinese translation in this work, and future work could extend the data and the method to other translation directions.
(2) There is still a lack of accurate automatic evaluation metrics for literary translation.
Previous literature translation work typically relies on human evaluation~\cite{fonteyne-etal-2020-literary,karpinska-iyyer-2023-large}, and points out that BLEU and Comet might not be suitable for evaluating literature translation~\cite{karpinska-iyyer-2023-large}.
This is because literary translations carry the responsibility of both semantic and critical interpretation, as they must address the challenge of achieving equivalence that often extends beyond the level of individual words~\cite{thai-etal-2022-exploring}.

\section*{Ethical Considerations}

We discuss the main ethical considerations of DRT models as follows: (1) Copyright. We mine literature sentences from 400 English books provided by the Project Gutenberg public-domain book repository\footnote{\url{https://www.gutenberg.org/}}, where the books are typically more than fifty years old and their copyrights have expired.
The book data also has been extracted and released by \citet{kryscinski-etal-2022-booksum}.
Therefore, we can construct DRT data based on these books, and further release our synthesized data.
(2) Licenses. We will release our model checkpoints and synthesized data under CC-BY-NC-SA 4.0 license.


\bibliography{custom}

\appendix

\clearpage

\section{Prompt in Data Synthesis}

\subsection{Prompts in Literature Book Mining}
\label{appendix:prompts_in_literture_book_mining}

\begin{tcolorbox}
\small

\texttt{SYSTEM PROMPT:} \\

You are assigned to translate an English literary work into Chinese. The text may include descriptions or expressions that embody English cultural nuances, which may not resonate with Chinese language habits. In such instances, a literal translation may not be appropriate; instead, these sentences should be paraphrased to convey their intended meaning effectively. \\

\texttt{USER PROMPT:} \\

The English sentence is provided as follows:\\
<english sentence>\\
\{sentence\}\\
</english sentence>\\

Please begin by assessing whether the English sentence contains any metaphors or similes. If there are none, respond with "no metaphors and no similes."\\

If the English sentence does contain metaphors or similes, provide a literal translation of them, and then evaluate whether the literal translation is appropriate and easy for Chinese natives to understand.\\

If it is suitable, format your response as follows (two lines):\\
"your literal translation for metaphors/similes here (in Chinese)"\\
"suitable"\\

If it is unsuitable, please provide the reason for the unsuitability. Format your response as follows (three lines):\\
"your literal translation for metaphors/similes here (in Chinese)"\\
"unsuitable"\\
"reason for unsuitability here (in Chinese)"

\end{tcolorbox}

\subsection{Prompts in Multi-Agent Framework}
\label{appendix:prompts_in_multi_agent_framework}

\noindent \textbf{Translator Agent} (Word-level translation)

\begin{tcolorbox}
\small

Given an English sentence, identify the important words (usually nouns, verbs, technical terms, and named entities that require special attention in translation) and translate them into Chinese. Output the translations in JSON format, for example:\\

\{"EnglishWord1": "ChineseTranslation", "EnglishWord2": "ChineseTranslation"\}\\

The Chinese translations can be a single translation or multiple options as deemed appropriate.

\end{tcolorbox}

\vspace{0.5ex}
\noindent \textbf{Translator Agent}  (Preliminary translation)

\begin{tcolorbox}
\small

\texttt{SYSTEM PROMPT:} \\

Given an English sentence and a JSON object containing potential translations of important keywords, produce a Chinese literal translation of the entire sentence. Please directly output the Chinese translation without any descriptions. \\

\texttt{USER PROMPT:} \\

<English Sentence>\\
\{sentence\}\\
</English Sentence>\\
<Potential Keyword Translation>\\
\{keyword translation\}\\
</Potential Keyword Translation>

\end{tcolorbox}

\vspace{0.5ex}
\noindent \textbf{Translator Agent}  (Refinement translation)

In the refine loop, the translator agent receives the feedback of the previous translation, and then provides a new translation.
The prompt is a multi-turn dialogue between the translator and advisor, where the system prompt is the same as the preliminary translation.

\vspace{0.5ex}
\noindent \textbf{Advisor Agent}

\begin{tcolorbox}
\small

Please rate the Chinese translation of the following English text and provide your comments and suggestions.

\end{tcolorbox}

\vspace{0.5ex}
\noindent \textbf{Evaluator Agent}

\begin{tcolorbox}
\small

\texttt{SYSTEM PROMPT:} \\

Please evaluate the following Chinese translation of an English text. Rate the translation on a scale of 0 to 100, where:\\
- 10 points: Poor translation; the text is somewhat understandable but contains significant errors and awkward phrasing that greatly hinder comprehension for a Chinese reader.\\
- 30 points: Fair translation; the text conveys the basic meaning but lacks fluency and contains several awkward phrases or inaccuracies, making it challenging for a Chinese reader to fully grasp the intended message.\\
- 50 points: Good translation; the text is mostly fluent and conveys the original meaning well, but may have minor awkwardness or slight inaccuracies that could confuse a Chinese reader.\\
- 70 points: Very good translation; the text is smooth and natural, effectively conveying the intended meaning, but may still have minor issues that could slightly affect understanding for a Chinese reader.\\
- 90 points: Excellent translation; the text is fluent and natural, conveying the original meaning clearly and effectively, with no significant issues that would hinder understanding for a Chinese reader.\\

Please provide the reason first, followed by a score. Format your evaluation in the JSON structure below:\\
\{"reason": "reason for the score", "score": int\}

\end{tcolorbox}

\subsection{Prompts in Thought Reformulation}
\label{appendix:prompts_in_thought_reformulation}

\begin{tcolorbox}
\small

A student is engaged in the task of translating an English sentence into Chinese.\\

The English sentence is as follows:\\
<English Sentence>\\
\{sentence\}\\
</English Sentence>\\

This student constantly thinks about and optimizes his translation. The whole process is shown as follows:\\

<Translation Process>\\
\{translation process\}\\
</Translation Process>\\

Please polish the whole translation process into a long first-person self-reflection description (use the present tense).\\

The self-reflection should begin with selecting the keywords from the English sentence, translating the keywords, and then attempt to translate the whole sentence, and then think about whether the translation is good or not, and iteratively make translation attempts. Finally, make a final translation decision.\\

Output the self-reflection description directly without any additional descriptions or explanations. Each line in the self-reflection description can be regarded as a reasoning step to the translation.

\end{tcolorbox}

\section{Details of Human Annotation}
\label{appendix:human_annotation}

In Section~\ref{subsec:2.4}, we employ human annotation to provide the quality comparison between two translations for a source sentence.
Specifically, we employ three Chinese master students with high levels of fluency in both English and Chinese as our human annotators.
For each sample, we give the source sentence and its two translation (without the scores provided by our evaluator agent) to all three annotators, and every annotator should provide one of the following judgments: (1) the first translation is better than the second one; (2) the second translation is better than the first one; (3) two translation are similar in quality.
During annotation, we encourage the annotators to give differentiated judgments, \emph{i.e.}, judgment (1) or judgment (2).
If three annotators give the same judgment for a sample, the judgment will be regarded as the final label.
Otherwise, its label will be decided by a group meeting of all three annotators and a senior data scientist.

\section{GPT-4o Evaluator}
\label{appendix:gpt_4o_evaluator}

For GRB and GRF, we prompt GPT-4o (2024-08-06 version) as the MT evaluator in the reference-based and reference-free manners, respectively.
The corresponding prompts borrow from \citet{kocmi2023large}, and make some adaptions to literature translation.

\vspace{0.5ex}
\noindent \textbf{GRB Prompt:}

\begin{tcolorbox}
\small

Score the following translation from English to Chinese with respect to the human reference on a continuous scale from 0 to 100, where score of zero means "no meaning preserved" and score of one hundred means "perfect preservation of meaning, with faithfulness, expressiveness, and elegance".\\

English source: \{src\}\\
Chinese human reference: \{ref\}\\
Chinese translation: \{hyp\}\\

Score:

\end{tcolorbox}

\vspace{0.5ex}
\noindent \textbf{GRF Prompt:}

\begin{tcolorbox}
\small

Score the following translation from English to Chinese on a continuous scale from 0 to 100, where score of zero means "no meaning preserved" and score of one hundred means "perfect preservation of meaning, with faithfulness, expressiveness, and elegance".\\

English source: \{src\}\\
Chinese translation: \{hyp\}\\

Score:

\end{tcolorbox}

\begin{table*}[t]
\centering
\resizebox{0.75\textwidth}{!}
{
\begin{tabular}{lcccccc}
\toprule[1pt]
\multicolumn{1}{c}{\multirow{2}{*}{Model}} & \multicolumn{3}{c}{reference-free} & \multicolumn{3}{c}{reference-based}                        \\
\cmidrule(r){2-4} \cmidrule(r){5-7} \multicolumn{1}{c}{}                       & GEA &  GRF               & CometKiwi             & GRB   & \multicolumn{1}{c}{BLEU} & \multicolumn{1}{c}{CometScore} \\ \midrule[1pt]
\multicolumn{7}{c}{\emph{Commercial LLMs}} \\ \midrule
GPT-4o &   71.88      &    85.57      &    \underline{73.01}      &    82.78      &    34.51      &    79.41  \\
o1-preview  &   \textbf{78.01}      &    \underline{87.11}      &    \textbf{73.70}      &    \textbf{83.86}      &    30.65      &    80.12  \\  \midrule[1pt]
\multicolumn{7}{c}{\emph{DRT}} \\ \midrule
DRT-8B $_\text{(Backbone: Llama-3.1-8B-Instruct)}$  & 69.65  & 84.49 & 70.85 & 80.80  & 32.67  & 78.81 \\ 
DRT-7B $_\text{(Backbone: Qwen2.5-7B-Instruct)}$  & 75.05  & 85.57 &  71.78    & 82.38  & \underline{35.54}   & \underline{80.19}    \\
DRT-14B $_\text{(Backbone: Qwen2.5-14B-Instruct)}$  & \underline{77.41}  & \textbf{87.19}  & 72.11   &  \underline{83.20} &   \textbf{36.46}     & \textbf{80.64}     \\ \bottomrule[1pt]
\end{tabular}
}
\caption{Experimental results of comparing DRT with commercial LLMs. The \textbf{bold} and the \underline{underline} denote the best and second-best performances, respectively.}
\label{table:commercial_res}
\end{table*}

\section{Implementation Details.}
\label{appendix:implementation_details}

\noindent \textbf{Automatic Evaluation.}
To calculate CometKiwi and CometScore, we leverage the official codes\footnote{\url{https://github.com/Unbabel/COMET}} and the official models\footnote{\url{https://huggingface.co/Unbabel/wmt22-cometkiwi-da} and \url{https://huggingface.co/Unbabel/wmt22-comet-da}}.
To calculate the BLEU score, we use the \textit{sacrebleu} toolkit\footnote{\url{https://github.com/mjpost/sacrebleu}} to calculate the corpus-level BLEU.

\vspace{0.2ex}
\noindent \textbf{Training Details.}
Llama-Factory~\cite{zheng-etal-2024-llamafactory} is used to instruct-tune LLMs.
All LLMs are tuned on 8$\times$NVIDIA A100 GPUs (40G) with 1e-5 learning rate and 8 (8$\times$1) batch size.
We use the DeepSpeed ZeRO-3 optimization~\cite{rasley2020deepspeed}.
Following~\citet{qin2024o1}, we set the number of training epochs to 3, and the training process costs 70 GPU hours and 124 GPU hours for 7B and 14B models, respectively.

\vspace{0.2ex}
\noindent \textbf{Inference Details.}
When evaluating model performance on the test set, we use vLLM toolkit~\cite{kwon2023efficient} to accelerate the model generation.
We use the sampling decoding strategy with 0.1 temperature, and set the repetition penalty to 1.05.
For DeepSeek-R1 series (DeepSeek-R1-Distill-Qwen-7B, DeepSeek-R1-Distill-Llama-8B, DeepSeek-R1-Distill-Qwen-14B and DeepSeek-R1-Distill-Qwen-32B), we follow the instruction\footnote{\url{https://github.com/deepseek-ai/DeepSeek-R1}} to enforce them to avoid blank thinking.
All experimental results listed in this paper are the average of 3 runs.

\section{Comparison with Commercial LLMs}
\label{appendix:commercial_llms}

As shown in Table~\ref{table:commercial_res}, DRT-14B achieves competitive results with o1-preview, showing its superiority.
Additionally, we observe that o1-preview significantly outperforms GPT-4o in terms of GEA. This finding highlights the effectiveness of long thought in machine translation. When applied to appropriate translation contexts, long thought can further enhance the authenticity of translations.



\end{document}